\pdfoutput=1

\documentclass[11pt]{article}

\usepackage[preprint]{acl}

\usepackage{times}
\usepackage{latexsym}
\usepackage{graphicx}

\usepackage[T1]{fontenc}

\usepackage[utf8]{inputenc}

\usepackage{microtype}

\usepackage{inconsolata}
\usepackage{amsmath}
\usepackage{amsfonts}
\usepackage[framemethod=TikZ]{mdframed}
\usepackage{tikz}
\usepackage{color,colortbl}
\usepackage{tcolorbox}
\usepackage{xspace}
\usepackage{algorithm}
\usepackage{multirow}
\usepackage{enumerate}
\usepackage{makecell}
\usepackage{arydshln}
\usepackage{pifont}
\usepackage{algpseudocode}
\usepackage{booktabs}
\usepackage{tablefootnote}
\usepackage{bbding}

\definecolor{blueish}{RGB}{250, 250, 255}
\definecolor{greenish}{RGB}{200, 255, 200}
\definecolor{ForestGreen}{RGB}{64, 136, 39}
\definecolor{redish}{RGB}{212, 57, 57}
\definecolor{highlight}{RGB}{175, 255, 100}
\definecolor{darkred}{RGB}{139, 0, 0}
\definecolor{gray95}{gray}{0.05}
\definecolor{rowgray}{RGB}{224, 224, 224}

\newmdenv[
    tikzsetting= {fill=blueish},
    skipabove=0.33em,
    skipbelow=0.33em,
    linewidth=1pt,
    innerleftmargin=4pt,
    innerrightmargin=4pt,
    innertopmargin=2pt,
    innerbottommargin=2pt,
    linecolor=gray95,
    roundcorner=2pt, 
    shadowsize=4pt,
    shadowcolor=gray95
]{answerbox}

\newenvironment{result}
{\begin{answerbox}}
{\end{answerbox}}

\newcommand{\CDD}{{\sc CDD}\xspace}
\newcommand{\CDDbf}{{\sc \textbf{CDD}}\xspace}
\newcommand{\TED}{{\sc TED}\xspace}
\newcommand{\TEDbf}{{\sc \textbf{TED}}\xspace}
\newcommand{\ndCDD}{{\sc DetCon}\xspace}
\newcommand{\ndCDDbf}{{\sc \textbf{DetCon}}\xspace}
\newcommand{\ndTED}{{\sc ComiEval}\xspace}
\newcommand{\ndTEDbf}{{\sc \textbf{ComiEval}}\xspace}

\title{Generalization or Memorization: Data Contamination and Trustworthy Evaluation for Large Language Models}

\author{\textbf{Yihong Dong},
    \textbf{Xue Jiang},
    \textbf{Huanyu Liu},
  \textbf{Zhi Jin}, 
  \textbf{Bin Gu}$^\dagger$, 
  \textbf{Mengfei Yang}$^\ddagger$,
   and \textbf{Ge Li}\footnotemark[1]\\
   Key Laboratory of High Confidence Software Technologies (Peking University), \\ Ministry of Education; School of Computer Science, Peking University, Beijing, China \\
   $^\dagger$Beijing Institute of Control Engineering, $^\ddagger$China Academy of Space Technology\\
    \texttt{\{dongyh, jiangxue\}@stu.pku.edu.cn}, 
    \texttt{\{zhijin, lige\}@pku.edu.cn} \\ 
}

\begin{document}
\maketitle
\renewcommand{\thefootnote}{\fnsymbol{footnote}}
\footnotetext[1]{Corresponding author.}
\renewcommand{\thefootnote}{\arabic{footnote}}

\begin{abstract}
Recent statements about the impressive capabilities of large language models (LLMs) are usually supported by evaluating on open-access benchmarks. Considering the vast size and wide-ranging sources of LLMs' training data, it could explicitly or implicitly include test data, leading to LLMs being more susceptible to data contamination. However, due to the opacity of training data, the black-box access of models, and the rapid growth of synthetic training data, detecting and mitigating data contamination for LLMs faces significant challenges. 
In this paper, we propose CDD, which stands for \textbf{C}ontamination \textbf{D}etection via output \textbf{D}istribution for LLMs. 
\CDD necessitates only the sampled texts to detect data contamination, by identifying the peakedness of LLM's output distribution. To mitigate the impact of data contamination in evaluation, we also present \TED: \textbf{T}rustworthy \textbf{E}valuation via output \textbf{D}istribution, based on the correction of LLM's output distribution. To facilitate this study, we introduce two benchmarks, i.e., \ndCDD and \ndTED, for data contamination detection and contamination mitigation evaluation tasks. Extensive experimental results show that 
\CDD achieves the average relative improvements of 21.8\%-30.2\% over other contamination detection approaches in terms of Accuracy, F1 Score, and AUC metrics, and can effectively detect implicit contamination. \TED substantially mitigates performance improvements up to 66.9\% attributed to data contamination across various contamination setups. In real-world applications, we reveal that ChatGPT exhibits a high potential to suffer from data contamination on HumanEval benchmark.\footnote{\url{https://github.com/YihongDong/CDD-TED4LLMs}}

\end{abstract}

\section{Introduction}
In recent years, LLMs have revolutionized the fields of natural language processing (NLP), artificial intelligence, and software engineering. To evaluate LLMs' capabilities in various downstream tasks, such as automatic question answering, natural language reasoning, and code generation, people conduct extensive tests for LLMs based on enormous benchmark datasets \cite{codex, cobbe2021gsm8k}. The results indicate that LLMs exhibit superior performance on these tasks. 
While marveling at the powerful capabilities of LLMs, people usually want to determine whether an LLM's excellent performance is due to the genuine understanding of tasks to achieve generalization, or merely because it has seen the test data to form memorization, i.e., suffering from data contamination.

\begin{figure}[t!]
    \centering
    \includegraphics[width=0.52\textwidth]{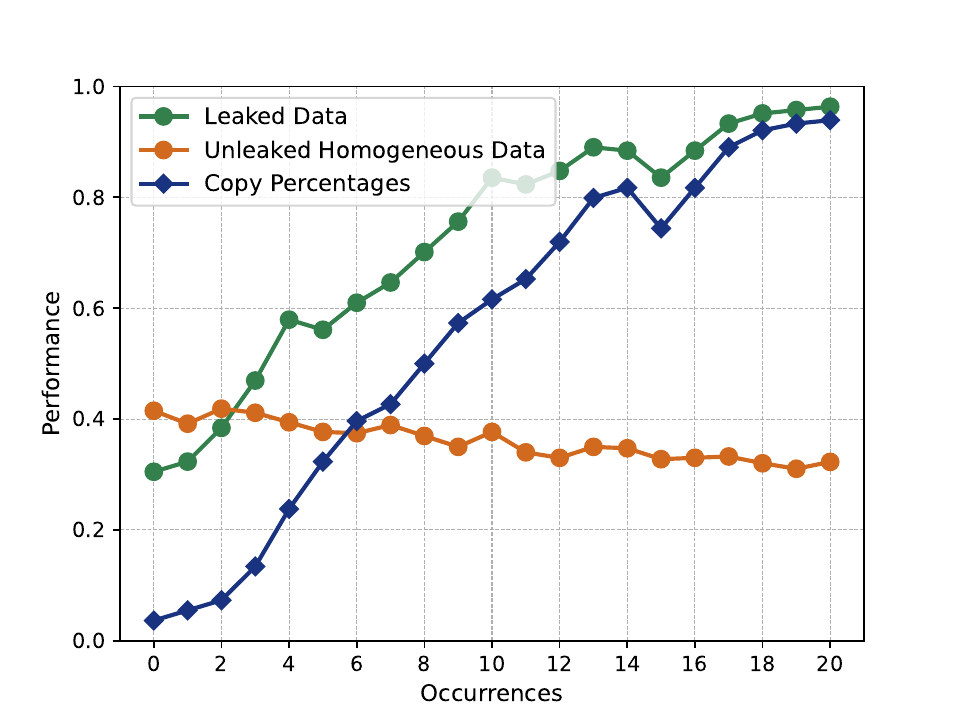}
    \caption{An example of data contamination affecting LLMs' performance, where CodeLlama is fine-tuned on HumanEval (as leaked data) + 50K StarCoder data excluding MBPP (as unleaked homogeneous data).}
    \label{fig1}
\end{figure}

Data contamination, also known as data leakage, refers to the scenario where the test data has been included in the model's training data \citep{DBLP:conf/acl/Magar022, bdtechtalks2023}, leading to the model performing exceptionally well on these leaked test data. Owing to the vast size and wide-ranging sources of the pre-trained datasets for LLMs, they are more susceptible to data contamination,  which can be primarily categorized into two situations: 1) For existing benchmark datasets, they are more easily leaked because of massive text quotes, code reuse, and synthetic data in LLMs' training data. 2) For upcoming benchmark datasets, newly constructed test data may already exist in the continuously evolving training data of LLMs since people are usually unaware of the specifics of LLMs' training data. Consequently, it becomes formidable to prevent data contamination for LLMs.

Data contamination exerts a profound and deleterious impact on LLMs \cite{cheater, stop, lensoftime}. 
As shown in Figure \ref{fig1}, with LLMs continuing to learn on contaminated data (i.e., both leaked data and other training data), their performance keeps improving on leaked data but stagnates and even degrades on similar data.
This example reflect that data contamination can lead to a substantial overestimation of models' performance, thus affecting the trustworthiness and effectiveness of LLMs in practical applications. Furthermore, data contamination may also conceal the potential flaws of models, presenting major obstacles for people to identify and improve upon LLMs' shortcomings. 
Therefore, it is crucial for LLMs to detect data contamination and ensure trustworthy evaluation.

Although acknowledged the significance, data contamination detection and trustworthy evaluation for LLMs still persist as open and challenging issues \cite{, Rephrased, Competition-Level-Coding}. The difficulties of data contamination detection can be essentially attributed to three factors: 
1) Opaque Training Data. It is usually non-public and comprehensive, while continuously evolving for new LLMs.
2) Black Box Models. The parameters and output probabilities of LLMs may not be available, such as ChatGPT and GPT-4 \cite{gpt-4}. 
3) Proliferation of Synthetic Data. It could implicitly introduce the variants\footnote{These variants may include, but are not limited to, translations into other languages, additions of explanations or intermediate processes, and provisions of alternate solutions.} of test data to training data. 
Further, the evaluation to mitigate the impact of data contamination has hardly been studied.

In this paper, we overcome the preceding challenges by proposing \CDD: \textbf{C}ontamination \textbf{D}etection via output \textbf{D}istribution for LLMs. \CDD uses the sampled texts to identify the peakedness of LLM's output distribution for data contamination detection. We follow a hypothesis that training is likely to alter the model's output distribution, resulting in a more peaked output distribution for training data, thereby tending the model towards specific outputs on these data. On this basis, we also present \TED: \textbf{T}rustworthy \textbf{E}valuation via output \textbf{D}istribution, which is designed to mitigate the impact of data contamination in evaluation by correcting LLM's output distribution. 

We construct two new datasets, i.e., \ndCDD and \ndTED, for data contamination detection and contamination mitigation evaluation tasks, respectively. Experimental results demonstrate that \CDD achieves state-of-the-art (SOTA) performance and is also suitable for identifying implicit contamination, i.e., existing the variants of test data in training data. \TED successfully mitigates the impact of data contamination in evaluation across various scenarios.
Furthermore, we also provide strong evidence that ChatGPT suffers from data contamination on HumanEval dataset.

\begin{figure}[t!]
    \centering
    \includegraphics[width=0.46\textwidth]{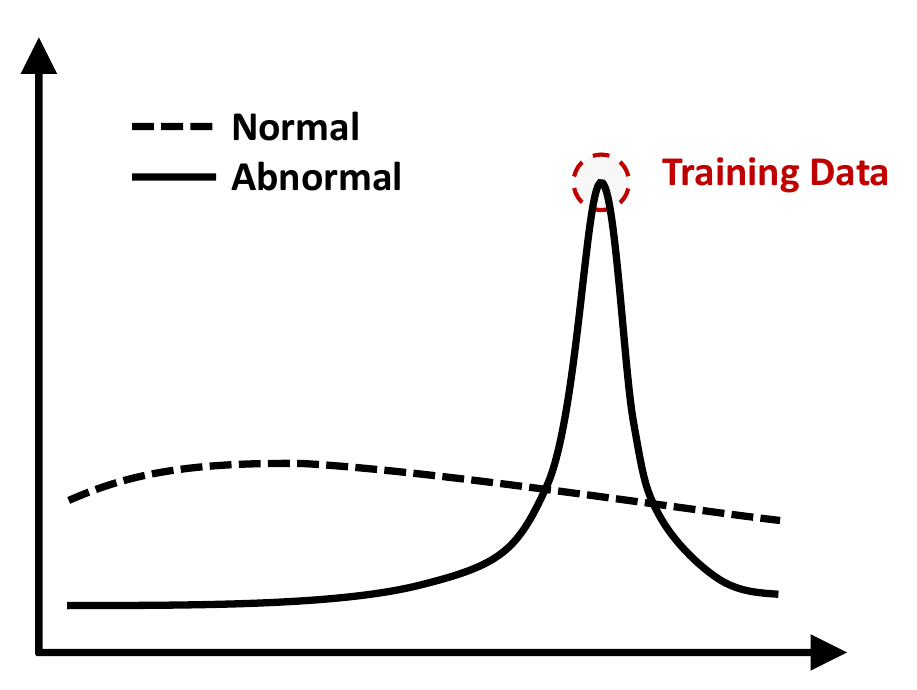}
    \caption{The illustration of LLMs' output distribution.}
    \label{fig2}
\end{figure}

\section{Motivation Example}
A powerful LLM that transcends memorization has the capability to generate diverse outputs in response to a given input. Considering the huge vocabulary size of LLMs, which encompasses a good number of tokens with analogous semantics, the output distribution sampled from LLMs ought to not exhibit peakedness. However, when LLMs solely form memorization via training, LLMs are prone to generate outputs that abnormally resemble their training data, as shown in Figure \ref{fig2}. From a statistical perspective, assuming that the average probability of LLM's output tokens is 0.95, the likelihood of sampling two outputs that contain the same 100 consecutive tokens is about 0.005 < 0.01, which is an extremely improbable event. Therefore, if an LLM consistently outputs some identical or highly similar texts through sampling, it is most likely caused by memorization.

\begin{figure}[t!]
    \centering
    \includegraphics[width=0.52\textwidth]{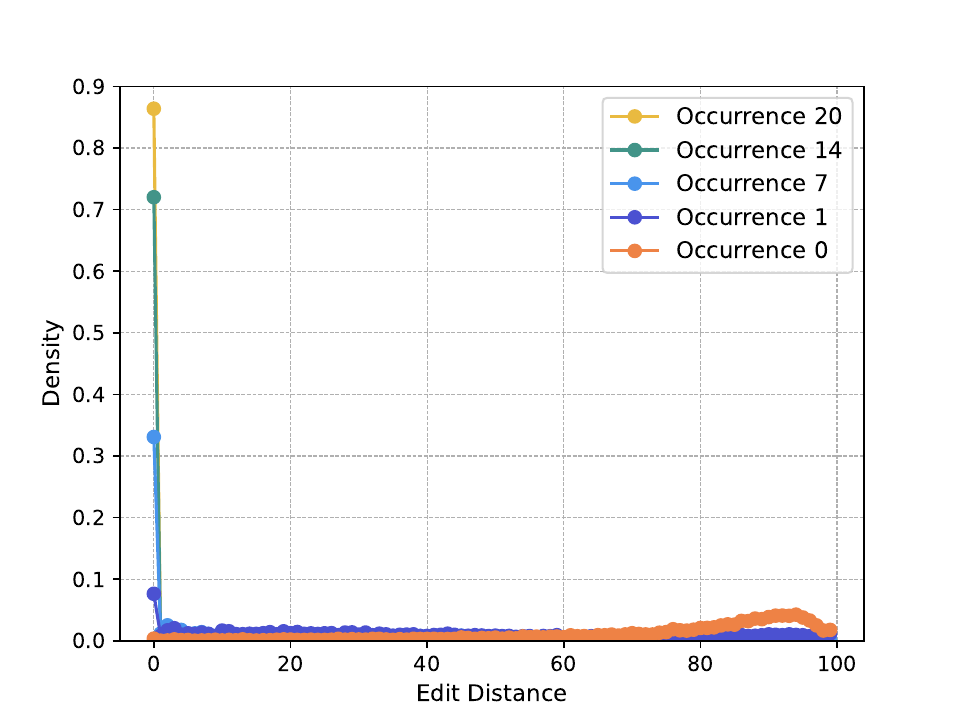}
    \caption{The output distributions of LLMs as modeled by edit distance across varying degrees of data contamination (with the same setting as Figure \ref{fig1}).}
    \label{fig3}
\end{figure}

Figure 3 displays an example of how the LLM's output distribution changes as the degree of data contamination varies. 
We model the LLM's output distribution by computing the edit distances of sampled texts, referred to as edit distance distribution ($\S~\ref{EDD}$).
As shown in Figure \ref{fig3}, in the absence of data contamination (i.e., occurrence 0), the density of zero edit distance stands at 0.0035, where zero edit distance means that sampled texts exactly match. 
However, upon the LLM being exposed to the leaked data even once (i.e., occurrence 1) during training, the density of zero edit distance escalates sharply to more than 20 times larger than the original, showing the peakedness. 
Therefore, the impact of data contamination on the LLM's output distribution is substantial.

In this paper, to the best of our knowledge, we are the first to consider from the standpoint of LLMs' output distribution to address the challenges associated with data contamination detection and contamination mitigation evaluation, employing only the sampled texts without access to the output probability and training data.

\section{Methodology}
In this section, we first establish the edit distance distribution ($\S~\ref{EDD}$), and then on this basis, we design CDD for data contamination detection ($\S~\ref{3.2}$) and TED for contamination mitigation evaluation ($\S~\ref{3.3}$). 

\subsection{Edit Distance Distribution}
\label{EDD}
Edit distance \cite{levenshtein1966binary} is a measure of similarity between two strings, which is defined as the minimum number of operations required to transform one string into the other. The operations typically include insertion, deletion, or substitution of a single character. 

Considering the generation of LLMs is based on tokens instead of characters, we adopt token-level edit distance in this paper. Given two strings $a$ and $b$, token-level edit distance is calculated as:
\begin{equation}
\small
\operatorname{ED}(a, b) = 
\begin{cases} 
\operatorname{Len}(a) \quad \text{if } \operatorname{Len}(b) = 0, \\
\operatorname{Len}(b) \quad \text{if } \operatorname{Len}(a) = 0, \\
\operatorname{ED}(\operatorname{Tail}(a), \operatorname{Tail}(b)) \quad \text{if } \operatorname{Head}(a) = \operatorname{Head}(b), \\
1 + \min\begin{cases}
\operatorname{ED}(\operatorname{Tail}(a), b) \\
\operatorname{ED}(a, \operatorname{Tail}(b)) \\
\operatorname{ED}(\operatorname{Tail}(a), \operatorname{Tail}(b))
\end{cases}  \text{otherwise,}
\end{cases}
\end{equation}
where $\operatorname{Len}(a)$ means the length of tokenized $a$, $\operatorname{Head}(a)$ means the first token of tokenized $a$, $\operatorname{Tail}(a)$ means the string consists of all tokens of tokenized $a$ following $\operatorname{Head}(a)$. 
We use dynamic programming to speed up calculations and rolling arrays to reduce space overhead.

Given an LLM, we can model its output distribution by computing the edit distances of sampled texts $S = \{s_1, s_2, ..., s_n\}$, where $n$ is the number of samples. Specifically, we define the density function $\rho$ as: 
\begin{equation}
    \rho(d) = \frac{\sum\limits_{i=1}^{n-1} \sum\limits_{j=i+1}^n \mathbb{I}(\operatorname{ED}(s_i, s_j) = d)}{n*(n-1)/2},
\end{equation}
where $d \in \mathbb{Z}_{\geq 0}$ and $\mathbb{I}(\cdot)$ is the indicator function that outputs 1 if the condition is true, otherwise 0. 

\subsection{\CDDbf for Data Contamination Detection}
\label{3.2}
Given a test data $\{x, y\}$ consisting of a prompt $x$ and the corresponding answer $y$, we aim to detect if this data has been trained by the model $\mathcal{M}$.

We sample $S$ from $\mathcal{M}$ with the input $x$ to calculate $\rho$. For data contamination detection, the calculation of $\rho$ can be simplified as:
\begin{equation}
    \rho'(d) = \frac{\sum\limits_{i=1}^n \mathbb{I}(\operatorname{ED}(s_i, y)=d)}{n}.
\end{equation}
However, $\rho'(d)$ assumes that test data must be explicitly leaked in its original form $\{x, y\}$, and does not take into account the possible implicit contamination of the variant form, i.e., $\{x, \hat{y}\}$. 

Through observation, we find that the copy percentage of model outputs increases as the degree of data contamination increases, as shown in Figure \ref{fig1}.  
Therefore, we approximate $y$ by the model's output texts and finally choose to replace $y$ with the model's greedy search text $s_{t=0}$, which can be easily achieved by setting temperature $t$ = 0 when sampling. Thus,
\begin{equation}
    \rho^*(d) = \frac{\sum\limits_{i=1}^n \mathbb{I}(\operatorname{ED}(s_i, s_{t=0})=d)}{n}.
\label{eq4}
\end{equation}
In this work, we employ $\rho^*(d)$ to measure edit distance distribution by default.

Further, we define the peakedness of edit distance distribution as
\begin{equation}
    \operatorname{Peak}(\mathcal{M}; x) = F(d \leq \alpha \cdot l) =\sum\limits_{d=0}^{\alpha \cdot l} \rho^*(d),
\label{eq5}
\end{equation}
where $F$ is the cumulative distribution function, $\alpha \in [0,1]$ is a hyper-parameter to control the similarity, and and $l$ is defined as:
\begin{equation}
    l = \max(\{\operatorname{Len}(s) \mid s \in S\}).
\label{eq7}
\end{equation}

Through identifying the peakedness, CDD can detect data contamination on test data as: 
\begin{equation}
\operatorname{CDD}(\mathcal{M}; x) = 
\begin{cases} 
\text{Leaked} \quad \text{if } \operatorname{Peak}(\mathcal{M}; x) > \xi, \\
\text{Unleaked} \quad \text{if } \operatorname{Peak}(\mathcal{M}; x) \leq \xi, 
\end{cases}
\label{eq6}
\end{equation}
where $\xi \in [0,1]$ is hyper-parameter to control the threshold.
The pseudocode of CDD for data contamination detection is shown in Algorithm \ref{algorithm 1}.

\begin{algorithm}
\caption{The pseudocode of CDD}
\label{algorithm 1}
\begin{algorithmic}[1]
\Require LLM $\mathcal{M}$, the prompt of test data $x$, and hyper-parameter $\alpha = 0.05, \xi = 0.01$.
\Ensure Contamination status $cs$.
    
    \State Sample $S$ from $\mathcal{M}$ with the input $x$.
    \State Model $\rho^*(d)$ via Eq. \eqref{eq4}
    \State Compute $\operatorname{Peak}(\mathcal{M}; x)$ via Eq. \eqref{eq5}.
    \State Detect $cs$ via Eq \eqref{eq6}
    \State \Return $cs$.
\end{algorithmic}
\end{algorithm}

\subsection{\TEDbf for Contamination Mitigation Evaluation}
\label{3.3}
We achieve contamination mitigation evaluation using TED, which includes two rules to correct the LLM's output distribution, i.e., exclude peakedness and remove duplicates.

1) Exclude Peakedness. We hope to restore the uncontaminated sampling results by excluding the peakedness in the LLM's output distribution, while excluding the greedy text $s_{t=0}$ which is most likely to represent the leaked data potentially memorized by the LLM.
\begin{equation}
    S_e = \{s \mid s \in S \land \operatorname{ED}(s, s_{t=0}) > \tau\},
\label{eq8}
\end{equation}
where $\tau \in [0, +\infty)$ is a hyper-parameter to control the difference.

2) Remove Duplicates. It aims to remove the duplicate sampling results, especially those differing from $s_{t=0}$, which are also less likely to duplicately occur in the uncontaminated sampling results.
\begin{equation}
S_r = \{s_i | s_i \in S \land \forall j < i, s_j \neq s_i\}.
\label{eq9}
\end{equation}

In the evaluation phase, an evaluation metric $\mathcal{E}$ using TED to mitigate the impact of data contamination can be defined as:
\begin{equation}
    \mathcal{E}_{\operatorname{TED}}(\mathcal{M}; x) \equiv \mathcal{E}_{\operatorname{TED}}(S; x) = \mathcal{E}(S_e \land S_r; x),
\label{eq10}
\end{equation}
The pseudocode of TED for contamination mitigation evaluation is shown in Algorithm \ref{algorithm 2}. 

\begin{algorithm}
\caption{The pseudocode of TED.}
\label{algorithm 2}
\begin{algorithmic}[1]
\Require LLM $\mathcal{M}$, the prompt of test data $x$, evaluation metric $\mathcal{E}$, and hyper-parameter $\tau = 2$.
\Ensure Evaluation performance $ep$.
\State Sample $S$ from $\mathcal{M}$ with the input $x$.
\State Exclude peakedness to compute $S_e$ via Eq. \eqref{eq8}.
\State Remove duplicates to compute $S_r$ via Eq. \eqref{eq9}.
\State Obtain $ep$ based on $\mathcal{E}$ via Eq. \eqref{eq10}.
\State \Return $ep$.
\end{algorithmic}
\end{algorithm}

\section{Experiment}
In this section, we first introduce two datasets, \ndCDD and \ndTED, tailored for the tasks of data contamination detection and contamination mitigation evaluation, respectively ($\S~\ref{Dataset}$). We then evaluate the efficacy of \CDD on \ndCDD dataset ($\S~\ref{DCD}$). Following this, we assess the performance of TED on \ndTED dataset ($\S~\ref{CME}$). Finally, we demonstrate the application results of both \CDD and \TED in real-world scenarios ($\S~\ref{RWA}$).

\begin{table*}[th!]
\centering
\caption{Detailed statistics of simulating different data contamination scenarios of LLMs.}
\label{table1}
\resizebox{1.01\textwidth}{!}{
\begin{tabular}{@{}lccccccc@{}}
\toprule
Domain  & \multicolumn{1}{c}{Leaked Dataset} & \multicolumn{1}{c}{Base LLMs} & \multicolumn{1}{c}{Other Training Data} & \multicolumn{1}{c}{Mixing Ratio}      & \multicolumn{1}{c}{Learning Rate}     & Occurrences                  & \multicolumn{1}{c}{Contamination Form} \\ \midrule
Code Generation   & HumanEval                       & \{CodeLlama, CodeGen\} & StarCoder data                          & \multirow{2}{*}{1 : \{0, 0.1K, 1K, 10K\}} & \multirow{2}{*}{\{1e-3, 2e-4, 4e-8\}} & \multirow{2}{*}{{[}0, 20{]}} & \multirow{2}{*}{\{Explicit, Implicit\footnote{}\}} \\
Logical Reasoning & GSM8K                           & \{Llama2, Bloom\} & RedPajama data                          &                                      &                                       &                              &       \\ \bottomrule
\end{tabular}}
\end{table*}
\footnotetext{We rephrase leaked data and each problem in the variant of leaked data has another correct solution different from the original solution, where the majority is generated by ChatGPT and about 10\% is generated by ChatGPT-assisted humans.}

\begin{table*}[th!]
\centering
\caption{The differences between \CDD and other contamination detection approaches, where N-gram and LLM Decontaminator are designed to detect the contamination of training data rather than models.}
\label{table3}
\resizebox{1.01\textwidth}{!}{
\begin{tabular}{@{}lcccc@{}}
\toprule
Approach              & \multicolumn{1}{c}{Not Need Prob.} & \multicolumn{1}{c}{Not Need Param.} & \multicolumn{1}{c}{Not Need Other LLM} & \multicolumn{1}{c}{Consider Implicit Contamination} \\ \midrule
N-gram \cite{gpt3}               & \textcolor{ForestGreen}{\CheckmarkBold}                                       & \textcolor{ForestGreen}{\CheckmarkBold}                                     & \textcolor{ForestGreen}{\CheckmarkBold}                                     & \textcolor{redish}{\XSolidBrush}                                                  \\
Embedding similarity  & \textcolor{ForestGreen}{\CheckmarkBold}                                       & \textcolor{redish}{\XSolidBrush}                                      & \textcolor{ForestGreen}{\CheckmarkBold}                                     & \textcolor{redish}{\XSolidBrush}                                                  \\
Perplexity \cite{ppl}          & \textcolor{redish}{\XSolidBrush}                                        & \textcolor{ForestGreen}{\CheckmarkBold}                                     & \textcolor{ForestGreen}{\CheckmarkBold}                                     & \textcolor{redish}{\XSolidBrush}                                                  \\
Min-k\% Prob \cite{min-k}         & \textcolor{redish}{\XSolidBrush}                                        & \textcolor{ForestGreen}{\CheckmarkBold}                                     & \textcolor{ForestGreen}{\CheckmarkBold}                                     & \textcolor{redish}{\XSolidBrush}                                                  \\
LLM Decontaminator \cite{Rephrased}  & \textcolor{ForestGreen}{\CheckmarkBold}                                       & \textcolor{ForestGreen}{\CheckmarkBold}                                     & \textcolor{redish}{\XSolidBrush}                                      & \textcolor{ForestGreen}{\CheckmarkBold}                                                 \\
\CDDbf & \textcolor{ForestGreen}{\CheckmarkBold}                                       & \textcolor{ForestGreen}{\CheckmarkBold}                                     & \textcolor{ForestGreen}{\CheckmarkBold}                                     & \textcolor{ForestGreen}{\CheckmarkBold}                                                 \\ \bottomrule
\end{tabular}}
\end{table*}

\begin{table*}[ht!]
\centering
\caption{Comparison of \CDD and other contamination detection approaches, where $\dag$ denotes the application of the approach needs additional conditions as shown in Table \ref{table3} and the \textbf{\textit{bold italic}} indicates the highest value other than \CDD, which is also the baseline of the relative improvement.}
\label{tabel4}
\resizebox{1.01\textwidth}{!}{
\begin{tabular}{@{}lcccccccc@{}}
\toprule
\multirow{2}{*}{Approach} & \multicolumn{3}{c}{\ndCDD (Code Generation)}        & \multicolumn{1}{c}{\multirow{2}{*}{Average}}                  & \multicolumn{3}{c}{\ndCDD (Logical   Reasoning)}                       & \multicolumn{1}{c}{\multirow{2}{*}{Average}} \\ \cmidrule(r){2-4} \cmidrule(r){6-8}
\multicolumn{1}{c}{}                          & \multicolumn{1}{l}{Accuracy} & \multicolumn{1}{l}{F1 Score} & \multicolumn{1}{l}{AUC} & & \multicolumn{1}{l}{Accuracy} & \multicolumn{1}{l}{F1 Score} & \multicolumn{1}{l}{AUC} &                          \\ \midrule
N-gram (char-level)                           & 0.484                        & 0.593                        & -                       & 0.538                                        & 0.564                        & 0.67                         & -                                  & 0.617                                        \\
N-gram (token-level)                          & 0.541                        & 0.302                        & -                       & 0.422                                        & 0.656                        & 0.498                        & -                                  & 0.577                                        \\
Embedding similarity$^\dag$                         & 0.524                        & 0.569                        & 0.571                   & 0.554                                        & 0.592                        & 0.645                       & 0.668                              & \textbf{\textit{0.635}}                                        \\
Perplexity$^\dag$                                   & 0.513                        & 0.593                        & 0.491                   & 0.532                                        & 0.497                        & 0.664                        & 0.699                     & 0.620                                        \\
Min-k\% Prob$^\dag$                                 & 0.563                        & 0.524                        & 0.565                   & 0.550                                        & 0.527                        & 0.677                        & 0.698                     & 0.634                                        \\
LLM Decontaminator$^\dag$                            & 0.535                        & 0.578                        & -                       & \textbf{\textit{0.556}}                                        & 0.509                        & 0.433                        & -                         & 0.471                                        \\
\CDDbf                        & \textbf{0.715}               & \textbf{0.694}               & \textbf{0.761}          & \textbf{0.724~(\textcolor{ForestGreen}{ $\uparrow$ 30.2\%})}                              & \textbf{0.706}               & \textbf{0.765}               & \textbf{0.846} & \textbf{0.773~(\textcolor{ForestGreen}{ $\uparrow$ 21.8\%})}                                                                    \\                               \bottomrule
\end{tabular}}
\end{table*}

\subsection{Dataset}
\label{Dataset}
Considering the absence of datasets for data contamination detection and contamination mitigation evaluation tasks, we dedicate more than 2100 hours to constructing the \ndCDD and \ndTED datasets, utilizing two A6000 GPUs (48GB $\times$ 2).

We simulate data contamination by training LLMs using benchmark data.  To cover various scenarios of data contamination, we consider different settings, including two domain benchmarks leaked on four LLMs, two contamination form (i.e. explicit and implicit leaked data), using three different learning rates during training, four mixing ratios of leaked data with other training data, and 21 degrees of contamination (i.e., occurrences). The detailed statistics can be found in Table \ref{table1}. Due to the high cost of large-scale pre-training, we employ LoRA \cite{LORA} to fine-tune the base models on these various settings. On this basis, we construct the \ndCDD and \ndTED datasets.

\paragraph{\ndCDDbf} contains 2224 data contamination detection tasks, covering two domains (code generation and logical reasoning) and two contamination forms (explicit and implicit), which need to detect whether a specific LLM has contamination on a particular data. We randomly select the data from the leaked dataset and the LLM from the settings in Table \ref{table1}, where occurrence 0 refers to `uncontaminated' and the others denote `contaminated'. 

\paragraph{\ndTEDbf} contains 560 contamination mitigation evaluation tasks, consisting of a randomly selected contaminated model from Table \ref{table1} and the corresponding uncontaminated model, which need to evaluate the performance of the contaminated model and try to mitigate the impact of data contamination to approach the performance of the uncontaminated model.

The detailed statistics and introductions of \ndCDD and \ndTED datasets can be found in Appendix \ref{appendix_dataset_construct}.

\begin{figure*}[th!]
    \centering
    \includegraphics[width=0.98\textwidth]{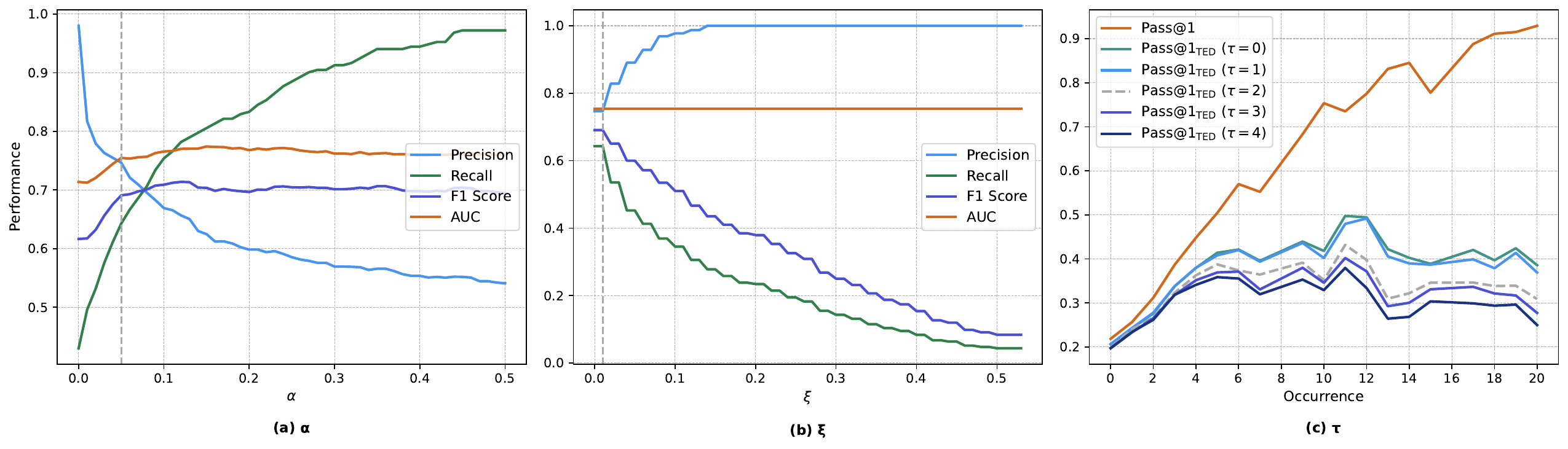}
    \caption{The influence of hyper-parameters, where $\alpha$ and $\xi$ serve for \CDD, $\tau$ is used for \TED, and we use the \textcolor{gray}{gray dashed line} to represent the employed hyper-parameters. }
    \label{fig4}
\end{figure*}

\begin{figure*}[th!]
    \centering
    \includegraphics[width=0.98\textwidth]{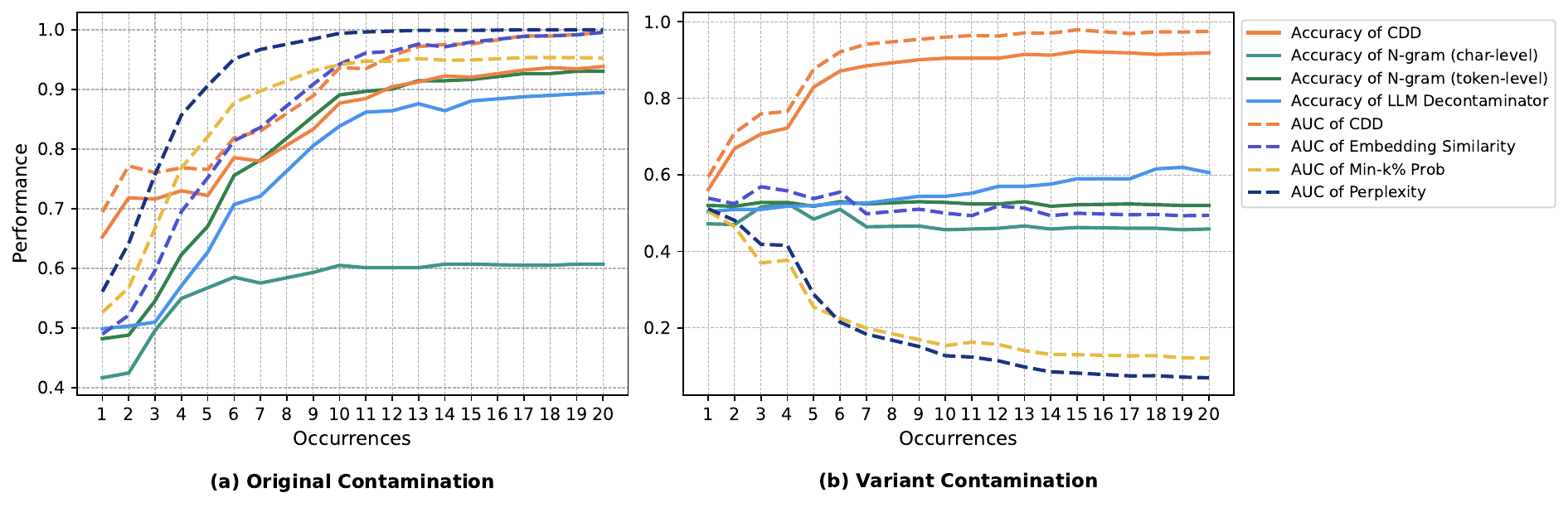}
    \caption{The effectiveness of \CDD for data contamination detection in explicit and implicit contamination forms.}
    \label{fig5}
\end{figure*}

\subsection{Data Contamination Detection}
\label{DCD}
\textbf{Experimental Setup.} We compare \CDD with baselines, including 1) \textbf{N-gram}: We employ widely-used 13-gram for both char-level and token level; 2) \textbf{Embedding Similarity}: Use the embedding of the base model to compute similarity; 3) \textbf{Perplexity}: Compute the perplexity of the original answer given the prompt; 4) \textbf{Min-k\% Prob}: Compute the minimum k\% probability of the original answer given the prompt, and 5) \textbf{LLM Decontaminator}: Use other LLM to determinate the similarity and we employ ChatGPT as this LLM. 
The differences between \CDD and baselines are shown in Table \ref{table3}. For hyper-parameters, we set $\alpha = 0.05$, $\xi = 0.01$, the cap of $l$ as 100 for CDD by default, and baselines follow the settings in their paper. 

\begin{figure*}[th!]
    \centering
    \includegraphics[width=0.96\textwidth]{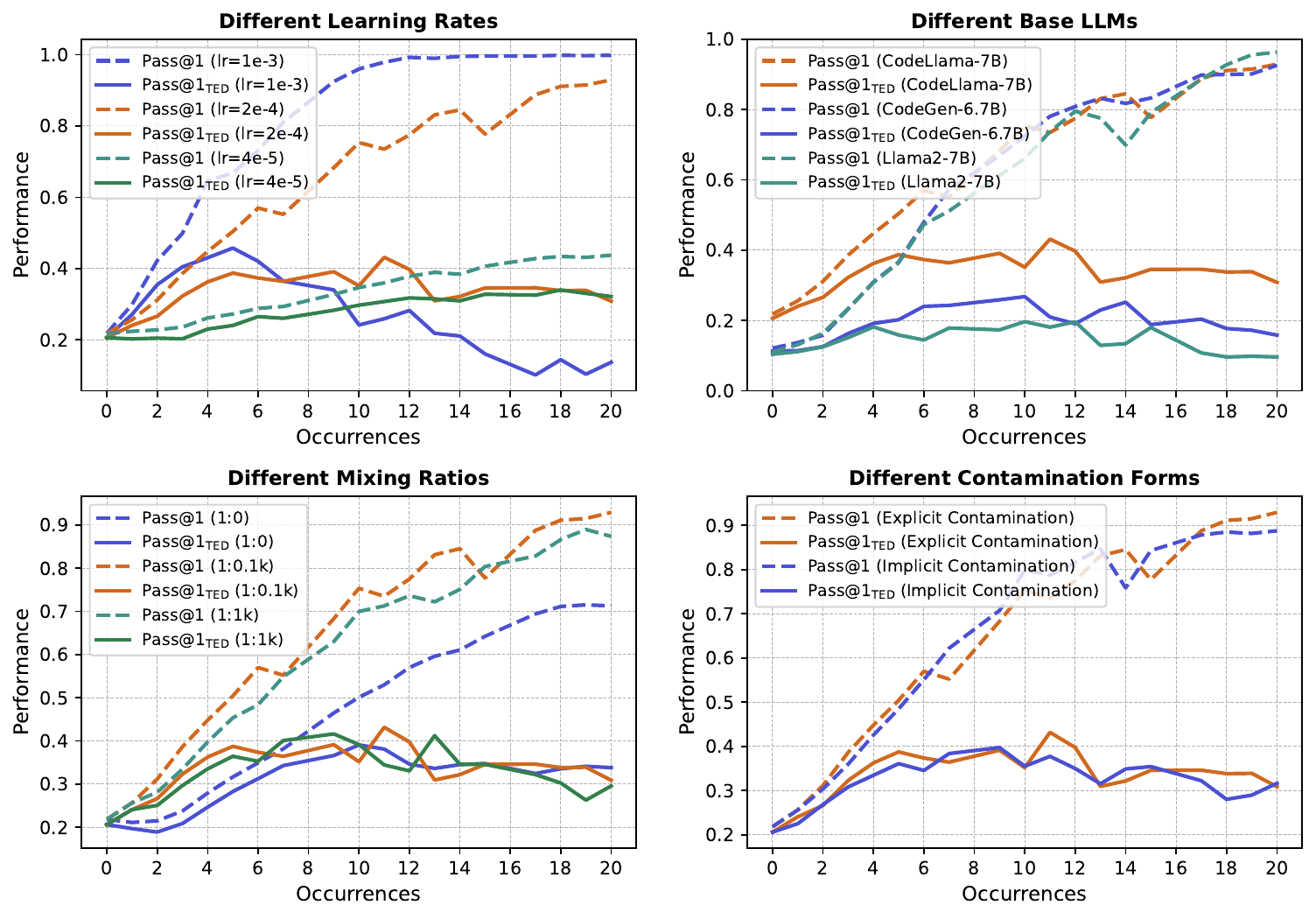}
    \caption{The effect of our TED on model performance as the degree of data contamination increases under different settings. The legend displays the settings for specific data leakage scenarios.}
    \label{occurrences}
\end{figure*}

\paragraph{The Effect of \CDDbf.} As presented in Table \ref{tabel4}, compared with other contamination detection approaches, \CDD attains SOTA performance in both code generation and logic reasoning domains. \CDD exhibits steady improvements across the Accuracy, F1 Score, and AUC metrics, with the average relative improvement ranging between 21.8\% and 30.2\%. Moreover, the advantage of \CDD is that it only requires the sampled texts of LLMs to detect data contamination, without the need for additional conditions in Table \ref{table3}.

We also evaluate the performance of  \CDD and other contamination detection approaches in identifying explicit and implicit forms of data contamination, as shown in Figure \ref{fig5}. In the cases of explicit contamination, as the degree of contamination increases, the detection effectiveness across all approaches improves. \CDD outperforms the other approaches at lower contamination degrees, which are more challenging to detect. In contrast, in the cases of implicit contamination, CDD alone maintains robust performance, whereas the other approaches encounter significant limitations. 

We fix the hyper-parameter $\alpha$ and $\xi$ intuitively for \CDD in the experiments. In Figure \ref{fig4} (a) and (b), we analyze the influence of $\alpha$ and $\xi$ empirically on \ndCDD dataset by changing itself and fixing another hyper-parameter. The results indicate that there is still room for further improvements with the better hyper-parameter setup of $\alpha$ and $\xi$.

\subsection{Contamination Mitigation Evaluation}
\label{CME}
\paragraph{Experimental Setup.} We evaluate the effectiveness of \TED for contamination mitigation in different learning rates, base LLMs, mixing ratios, contamination forms, and occurrences on \ndTED. We set the hyper-parameter $\tau = 2$ and use Pass@1 \cite{codex} as the evaluation metric $\mathcal{E}$.

\paragraph{Effect of \TEDbf.} \TED can steadily mitigate the performance improvements across different settings and occurrences in data contamination scenarios, as shown in Figure \ref{occurrences}. Moreover, the advantage of \TED is that the performance influence of \TED on the uncontaminated model (i.e. 0 occurrences) is small and almost negligible. However, as contamination degrees continue to increase, the performance influence of \TED becomes apparent in all of the different settings.

\begin{table}[h!]
\centering
\caption{Ablation Study of \TED, where RD and EP mean the rules of remove duplicates and exclude peakedness in \TED, respectively.}
\label{table5}
\resizebox{0.48\textwidth}{!}{
\begin{tabular}{@{}lccccc@{}}
\toprule
                              & \multicolumn{5}{c}{Occurrences}                            \\
\cmidrule(r){2-6}
\multirow{-2}{*}{Approach}    & 0 & 1      & 7      & 14     & 20     \\
\midrule
$\text{Pass@1}$ & 0.219                & 0.257 & 0.553 & 0.846 & 0.930 \\
+ RD     & 0.212                & 0.244 & 0.486 & 0.740 & 0.831~(\textbf{\textcolor{redish}{$\downarrow$ 10.7\%}}) \\
+ EP     & 0.212                & 0.242  & 0.371 & 0.335 & 0.320~(\textbf{\textcolor{redish}{$\downarrow$ 65.5\%}}) \\
$\text{Pass@1}_{\text{\TED}}$ & 0.209  & 0.241 & 0.364 & 0.321 & 0.308~(\textbf{\textcolor{redish}{$\downarrow$ 66.9\%}})\\
\bottomrule
\end{tabular}}
\end{table}

We analyze the effects of each component in \TED, as shown in Table \ref{table5}. The main function is provided by the rule of exclude peakedness, followed by the rule of remove duplicates. Both components are beneficial to \TED and are also effective when employed alone.

As illustrated in Figure \ref{fig4} (c), an increase in the hyperparameter $\tau$ for \TED leads to a more pronounced suppression of performance improvements attributable to data contamination. Meanwhile, it also marginally decreases the performance of the uncontaminated model.

\subsection{Real-World Application}
\label{RWA}
\paragraph{Experimental Setup.} 
In real-world applications, we apply \CDD and \TED for ChatGPT and construct two new datasets to assist evidence: 1) CodeForces2305 comprises 90 of the easiest level programming problems collected from the CodeForces website since May 2023, which is after the most recent update deadline of ChatGPT's training data, i.e., April 2023.  2) HumanEval\_R is reconstructed on HumanEval, which replaces its function signature, translates its requirements into German, French, and Chinese, selects different public test cases from the work \cite{CodeScore} to prompt, and remains the private test cases for testing. To enhance the detection precision, we set the hyper-parameters $\alpha$ to  0 and $\xi$ to a larger value of 0.2 for \CDD. We keep $\tau$ at the default value of 2 for \TED.

\begin{table}[th!]
\centering
\caption{Data contamination detection and contamination mitigation evaluation for ChatGPT, where `pre' and `post' present ChatGPT's APIs with fixed versions `0613' and '1106' respectively, Avg. Peak means the average of the peakedness of output distribution computed via Eq. \ref{eq5}, and CR means the ratio of contaminated tasks detected by \CDD in the benchmark.}
\label{tab:my-table}
\resizebox{0.48\textwidth}{!}{
\begin{tabular}{@{}lccccc@{}}
\toprule
Benchmark                        & Model & Pass@1 & Avg. Peak & CR & $\text{Pass@1}_{\text{\TED}}$ \\ \midrule
                                 & pre                          & 0.6131 & 0.1314                      & 0.2379     & 0.5535      \\
\multirow{-2}{*}{HumanEval}      & post                         & 0.7248 & 0.2326                      & 0.4147     & 0.5964      \\ \hdashline
                                 & pre                          & 0.4301 & 0.0455                      & 0.0976     & 0.4012      \\
                                 
\multirow{-2}{*}{HumanEval\_R}   & post                         & 0.4684 & 0.0594                      & 0.1097     & 0.4171      \\
\hdashline
                                 & pre                          & 0.0619 & 0.0049                      & 0          & 0.0616      \\
\multirow{-2}{*}{CodeForces2305} & post                         & 0.0790 & 0.0063                      & 0          & 0.0785      \\ \bottomrule
\end{tabular}}
\end{table}

\paragraph{Data contamination for ChatGPT.} 
As shown in Table \ref{tab:my-table}, on HumanEval dataset, both two versions of ChatGPT exhibit high Avg. Peak and Leak Ratio, and the `post' version become higher as ChatGPT continues to learn on new data. Considering the implementation of more stringent $\alpha$ and $\xi$, it is posited that ChatGPT is likely to suffer from data contamination on HumanEval dataset and become more serious over time.  This hypothesis is further evidenced through evaluations conducted on HumanEval\_R and CodeForces2305 datasets. HumanEval\_R indicates their high Avg. Peak and Leak Ratio are not easily attributable to the difficulty of problems. By modifying prompt forms through a process of reconstruction, all of the performance, Avg. Peak, and Leak Ratio of ChatGPT are significantly reduced. On CodeForces2305 dataset, which is unlikely to be involved in data contamination, ChatGPT's performance was markedly lower than anticipated, with the Avg. Peak at less than 0.01 and Leak Ratio of 0. Moreover, \TED demonstrates significant effectiveness on both the contaminated HumanEval and HumanEval\_R.


\section{Related Work}

\paragraph{Data contamination detection.} 
The concept of data contamination for LLMs can be derived from the context of GPT-3 \cite{gpt3}. Due to the vastness of the pre-training corpus of GPT-3, it inevitably overlapped with some evaluation benchmarks. Therefore, GPT-3 adopted 13-gram overlap detection to remove the data in the training set that conflicts with the test set of benchmarks. 

Some work \cite{DBLP:conf/sp/PanZJY20,cheater,stop,DBLP:conf/emnlp/DodgeSMAIGM021} exposed the serious consequences of data contamination and urged attention to this problem.
However, most currently released LLMs did not open their pre-training corpus, which poses a new challenge for data contamination detection.
Recent work tried to detect contamination without access to the pre-training corpus \cite{reranking, fillblank, timetravel}. Min-k\% Prob \cite{min-k} calculated the average of the k\% smallest probabilities of generated tokens and considered it as contaminated if it exceeded a certain threshold. The work \cite{ppl} assumed that data leaked into the training set tends to exhibit lower perplexity and utilizes perplexity analysis for detection. However, they often require other model outputs (e.g. probability) in addition to text, presenting challenges in detecting closed-source LLMs like ChatGPT, and they ignore the potential implicit contamination from variants of test data.

Recent investigations \cite{Competition-Level-Coding, Rephrased} have suggested that filtering training data based on n-grams may not effectively address the issue of data contamination, especially concerning semantically equivalent sentence rephrasing. To this end, LLM Decontaminator \cite{Rephrased} detected the similarity of test data and training data based on other advanced LLMs.

Our work requires only sampled texts to detect LLM's data contamination via output distribution and considers the potential implicit contamination. 

\paragraph{Contamination Mitigation Evaluation.}

To mitigate the impact of data contamination and ensure trustworthy evaluations, several approaches focus on constructing new evaluation benchmarks \cite{timetravel}. The work \cite{cleaneval} employs an LLM to paraphrase the contaminated dataset for evaluations. However, LLM's synthetic data is widely used for training, which already contains lots of paraphrased data \cite{Rephrased}. The work \cite{latesteval} leverages temporal information to construct a benchmark beginning from January 2023. However, building a high-quality benchmark is costly and time-consuming, and unfortunately, the training data deadline for ChatGPT and GPT-4 has been updated from September 2021 to April 2023 and continues to be delayed.

Our work achieves contamination mitigation evaluation from the standard of LLM's output distribution and is orthogonal to the preceding works.


\section{Conclusion}
In this paper, we have proposed two novel approaches, namely \CDD and \TED, to deal with data contamination detection and contamination mitigation evaluation for LLMs, considering the LLM's output distribution. We construct two corresponding datasets, i.e., \ndCDD and \ndTED, for these two tasks. Extensive experimental results indicate the superiority and versatility of \CDD and \TED. Moreover, we also discover that ChatGPT is likely to suffer from data contamination on HumanEval dataset. 
We hope to shed light on this direction and call more attention to data contamination issues.


\section{Limitations}
Our work has several limitations, which we aim to address in our future work:

First, the validation of our work is mainly focused on benchmarks for code generation and logical reasoning, which are highly representative and widely adopted. In the future, we will further validate our approaches on other benchmarks.

Second, our approaches require multiple samplings to compute the output distribution, and the more samplings conducted, the better the effect. We can use parallel sampling techniques to speed up sampling, thereby reducing time overhead.

Third, considering the limitation of computational resources, we employ a popular parameter-efficient fine-tuning approach, i.e., LoRA, instead of full-parameter fine-tuning to simulate data contamination for LLMs. In future work, we plan to attempt full-parameter fine-tuning.

Finally, in constructing our datasets, we assume that the four base LLMs used do not suffer from data contamination on the selected benchmarks. However, in reality, these LLMs may have slight data contamination. To completely avoid this issue, it might be necessary to retrain an LLM from scratch on a training set known to be entirely free of test data. However, undertaking such a process would be prohibitively costly.

\section{Acknowledgments}
This research is supported by the National Natural Science Foundation of China under Grant No.62192730, 62192733, 61832009, 62192731, 62072007, the Key Program of Hubei under Grant JD2023008.

\bibliography{ref}
\clearpage

\newpage

\onecolumn

\begin{table*}[h!]
\centering
\caption{The statistics of \ndCDD and \ndTED datasets, where each text is equipped with the probability.}
\label{tabel2}
\resizebox{1.01\textwidth}{!}{
\begin{tabular}{@{}lccc@{}}
\toprule
Dataset               & Task Nums & Inputs (optional) per task                                                 & Outputs per task                  \\ \midrule
\ndCDD & 1112 / 1112      & a prompt, the original answer, 51 sampled texts, and the model parameter      & `contaminated' / `uncontaminated' \\
\ndTED & 560       & leaked dataset, 51 sampled texts of each leaked data, and model parameters & evaluation performance            \\ \bottomrule
\end{tabular}}
\end{table*}

\appendix

\section{Details of Dataset Construction}\label{appendix_dataset_construct}

In this section, we further describes the different data contamination scenarios, as well as how we collect and process data from these scenarios to construct the dataset.

First, we prepare the data and models:
\begin{enumerate}
    \item \textbf{Test Data.} We choose the HumanEval \cite{codex} dataset for code generation and the GSM8K \cite{cobbe2021gsm8k} dataset for logical reasoning.
    \item \textbf{LLMs.} For code generation tasks, we use CodeLlama-7B \cite{codellama} and CodeGen-6.7B \cite{CodeGen}; for logical reasoning tasks, we select Llama2-7B \cite{Llama2} and Bloom-7B \cite{BLOOM}.
    \item \textbf{Training Data.} Code generation tasks use the training data from StarCoder \cite{StarCoder}, while logical reasoning tasks use RedPajama \cite{Redpajama}.
\end{enumerate}

Next, we construct the dataset \ndCDD for data contamination detection, starting with the construction of uncontaminated samples. We directly use the outputs generated by LLMs on the test data, representing uncontaminated data. Then, we construct contaminated samples by simulating different contamination scenarios:

\begin{enumerate}
\item \textbf{Explicit and Implicit Contamination}. Explicit contamination refers to the direct use of test data for training, while implicit contamination refers to training with variants of the test data.
\item \textbf{Proportion in the training data}. We use different amounts of training data mixed with test data to train LLMs. The proportions of the test dataset mixed with training data include 1:0, 1:0.1k, 1:1k, 1:10k.
\item \textbf{Different learning rates}. Considering the effect of learning rate on model training, we chose three different learning rates: 1e-3, 2e-4, and 4e-8.
\item \textbf{Degree of data contamination}. Training LLMs with contaminated data for more epochs indicates a higher degree of contamination. Epochs range from 0 to 20, where 0 means no training of LLM, indicating no contamination, reserved for constructing uncontaminated samples.

\end{enumerate}

By combining these four different scenarios, we can construct a variety of composite data contamination scenarios. For each piece of test data, we randomly select the results generated by LLMs under one of the contamination scenarios as the contaminated samples. Following the previous works \cite{Self-planning, Self-collaboration, CODEP, DBLP:journals/corr/abs-2401-06401}\nocite{Subtoken-TranX}, in generating these samples, we also record the outputs of greedy search with a temperature parameter of 0 (1 sample) and 50 samples obtained by sampling with a temperature of 0.8.

This construction approach aims to comprehensively cover possible data contamination scenarios, ensuring we can accurately assess the performance of LLMs in the face of different types and degrees of data contamination.

Finally, we construct the dataset \ndTED for contamination mitigation evaluation. We selected 560 LLMs and their generated outputs from all the constructed contaminated LLMs as the task inputs. Then, we used the performance of the corresponding uncontaminated LLMs on these two test datasets as the target output, serving as the evaluation criterion. Table \ref{tabel2} demonstrates the statistics of \ndCDD and \ndTED, respectively.

\section{Case Study}
We display the first 10 samples and the sample with greedy search is marked in bold as shown below.

\begin{figure}[h!]
    \centering
    \includegraphics[width=0.91\textwidth]{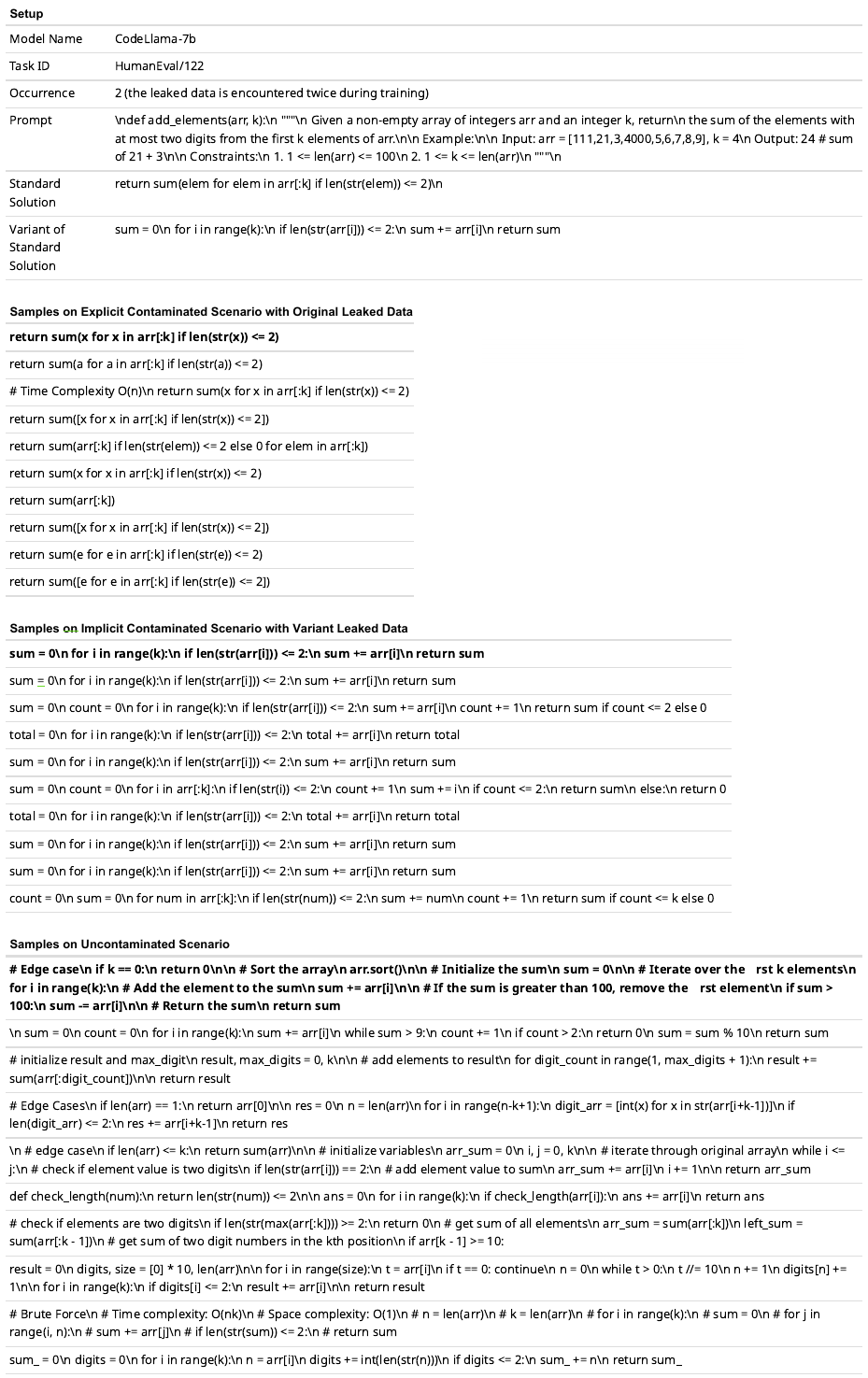}
\end{figure}

\end{document}